\documentclass[pdflatex,sn-mathphys]{sn-jnl}
\usepackage{amsmath}
\usepackage{amssymb}
\usepackage{xfrac}
\usepackage{nicefrac}
\usepackage{graphicx}
\usepackage{epstopdf}
\usepackage{enumerate}
\usepackage{subfig}
\PassOptionsToPackage{hyphens}{url}\usepackage{hyperref} \hypersetup{colorlinks,urlcolor=blue}
\usepackage{comment}
\usepackage{relsize}
\usepackage{multirow}
\usepackage{makecell}
\usepackage{footnote}
\usepackage[flushleft]{threeparttable}
\usepackage{caption}
\DeclareCaptionLabelSeparator{none}{ }
\makeatletter
\def\@acknow{}%
\long\def\EarlyAcknow#1 \par{%
\def\@acknow{\abstractfont\abstracthead*{Acknowledgments}% or use \subabstracthead <<<<
#1\par}}%
\def\printabstract{\ifx\@abstract\empty\else\@abstract\fi\par%
                   \ifx\@acknow\empty\else\@acknow\fi\par}
\makeatother

\jyear{2022}%
\begin{document}

\title[A Survey of Robotic Harvesting Systems and Enabling Technologies]{A Survey of Robotic Harvesting Systems and Enabling Technologies}

\author*[1]{\fnm{Leonidas}    \sur{Droukas}}      \email{ldroukas@ece.auth.gr}
\author[1]{ \fnm{Zoe}         \sur{Doulgeri}}     \email{doulgeri@ece.auth.gr}
\author[1]{ \fnm{Nikolaos L.} \sur{Tsakiridis}}   \email{tsakirin@ece.auth.gr}
\author[2]{ \fnm{Dimitra}     \sur{Triantafyllou}}\email{dtriant@iti.gr}
\author[2]{ \fnm{Ioannis}     \sur{Kleitsiotis}}  \email{ioklei@iti.gr}
\author[2]{ \fnm{Ioannis}     \sur{Mariolis}}     \email{ymariolis@iti.gr}
\author[2]{ \fnm{Dimitrios}   \sur{Giakoumis}}    \email{dgiakoum@iti.gr}
\author[2]{ \fnm{Dimitrios}   \sur{Tzovaras}}     \email{Dimitrios.Tzovaras@iti.gr}
\author[3]{ \fnm{Dimitrios}   \sur{Kateris}}      \email{d.kateris@certh.gr}
\author[3]{ \fnm{Dionysis}    \sur{Bochtis}}      \email{d.bochtis@certh.gr}

\affil[1]{\orgdiv{Department of Electrical and Computer Engineering}, \orgname{Aristotle University of Thessaloniki (AUTH)}, \orgaddress{\city{Thessaloniki}, \postcode{54124}, \country{Greece}}}
\affil[2]{\orgdiv{Information Technologies Institute (ITI)}, \orgname{Centre for Research and Technology Hellas (CERTH)}, \orgaddress{\city{Thessaloniki}, \postcode{57001}, \country{Greece}}}
\affil[3]{\orgdiv{Institute for Bio-Economy and Agri-Technology (iBO)}, \orgname{Centre for Research and Technology Hellas (CERTH)}, \orgaddress{\city{Volos}, \postcode{38333}, \country{Greece}}}

%-----------------------------------------------------------------------------------------------------------------------------------------------------

\abstract{
This paper presents a comprehensive review of ground agricultural robotic systems and applications with special focus on harvesting that span research and commercial products and results, as well as their enabling technologies. The majority of literature concerns the development of crop detection, field navigation via vision and their related challenges. Health monitoring, yield estimation, water status inspection, seed planting and weed removal are frequently encountered tasks. Regarding robotic harvesting, apples, strawberries, tomatoes and sweet peppers are mainly the crops considered in publications, research projects and commercial products. The reported harvesting agricultural robotic solutions, typically consist of a mobile platform, a single robotic arm/manipulator and various navigation/vision systems. This paper reviews reported development of specific functionalities and hardware, typically required by an operating agricultural robot harvester; they include (a) vision systems, (b) motion planning/navigation methodologies (for the robotic platform and/or arm), (c) Human-Robot-Interaction (HRI) strategies with 3D visualization, (d) system operation planning \& grasping strategies and (e) robotic end-effector/gripper design. Clearly, automated agriculture and specifically autonomous harvesting via robotic systems is a research area that remains wide open, offering several challenges where new contributions can be made.
}

\EarlyAcknow{This research received funding from the European Community’s Framework Programme Horizon 2020 under grant agreement No 871704, project BACCHUS.
}

     
\maketitle

%-----------------------------------------------------------------------------------------------------------------------------------------------------

\section{Introduction}  \label{intro}

The growing demand for food supply that derives from the continuously increasing population\footnote{\href{https://population.un.org/wpp/Graphs/Probabilistic/POP/TOT/900}{https://population.un.org/wpp/Graphs/Probabilistic/POP/TOT/900} - Last accessed: 22-11-2022} has made agricultural productivity growth an important priority. Labor availability pressure driven by demographics of an aging population, increasing urbanization, climate change and land degradation, as well as certain limitations regarding the arable land availability push forward slowly but steadily, the use of advanced agricultural technologies. Incorporating such technologies into agricultural production benefits the overall productivity and in turn supports the economic development and growth \cite{Eberhardt}. Additionally, automation in agriculture is bound to help improve the difficult work conditions of farmers and agricultural workers that are generally linked to various musculoskeletal disorders \cite{Fathallah,Proto,zZhang}. Application of robotic solutions regarding crop monitoring and harvesting is considered to have significant beneficial effects on production profits \cite{zZhang2}, enabling faster and easier automated harvest and increasing crop quality and yield. Thus, the development of robotic technologies and their application in agriculture is becoming a growing topic of interest and consideration \cite{Marinoudi}, with an increasing amount of research work being noticed in the last decades. This falls under the general umbrella of a growing trend in agriculture, which is termed \textit{precision agriculture} or agriculture with reduced carbon footprint.

Several reviews may be found in the literature, focusing on agricultural robots \cite{Fountas,Ramin,Bac,Oliveira}. Standardisation of terms and characteristics and system performance measures are utilized to evaluate these robotic systems for various field tasks and operations, e.g. transplanting/seeding, pruning/thinning, weed control and disease monitoring \cite{Slaughter,qZhang,Bechar}. Focused research reviews have been conducted regarding technologies, dealing with specific issues in automated agriculture, such as the management and analysis of the large data sets that most robotic systems require during their operation \cite{Saiz} or the optimized sensing and detection required for visual-based guidance of the applied robotic agricultural systems \cite{Baillie}. Robotic applications that target specific crops have also been a research goal, including cotton \cite{Fue}, strawberries \cite{Defterli} or the general arable farming, e.g. wheat and rice \cite{Aravind}. A brief review concerning various apple harvesting platforms is presented in \cite{zZhang2} that additionally develops a software tool for their economic evaluation, while several literature studies on agricultural robotic systems are investigated including the AgROS emulation tool \cite{Tsolakis}, classifying them with regard to their analysis approach and farming operations.

In addition to UGVs i.e. Unmanned Ground Vehicles (robots), drones/Unmanned Aerial Vehicles (UAVs) have also been increasingly considered in agriculture during the last years \cite{Cerro}. Given their ability to easily cover large distances over the fields, aerial robots are widely utilized with respect to several agricultural tasks such as crop monitoring/mapping, pesticide/fertilizer spraying, seed sowing and growth assessment \cite{Mogili,Jeongeun,LeiFeng}. Combining the survey capabilities of UAVs with the targeted intervention ability of UGVs, multi-robot systems are gaining great attention as well \cite{ChanyoungJu,Ribeiro,Subodh}. Various works can be found in literature regarding the coordination of cooperative ground and aerial robots towards e.g. selective spraying \cite{Pretto}, soil and biota sampling \cite{Deusdado} and monitoring/inspection tasks \cite{Edmonds}.

Given the great amount of works corresponding to automated/precision agriculture, in this work we focus our investigation on ground agricultural robotic systems and applications. The presented review covers not only publications found in literature, but research projects and commercial products in the market as well, considering robotic solutions and their enabling technologies with a special focus on harvesting. Section 2 presents an overview of complete harvesting robotic systems for various types of crops. Section 3 investigates a variety of technologies typically required by an operating robot harvester including vision systems, navigation methodologies, HRI strategies and more. Sections 4 and 5 present past and existing research projects and commercial products respectively, followed by the Conclusion Section.

%-----------------------------------------------------------------------------------------------------------------------------------------------------

\section{Complete Harvesting Robotic Systems}  \label{ch2}

Robotic automated crop harvesting e.g. of fruits or vegetables has a high impact on agricultural productivity with the task being investigated and considered even from early 1960’s \cite{pLi}. There are two basic concepts regarding automated harvesting by robots: bulk and selective. Bulk concept involves the harvesting of all fruits without exceptions, usually considering methodologies such as tree trunk or limb shaking \cite{Coppock,Sumner,Torregrosa}. However, bulk methods entail the danger of harming the crops. With the development of new technologies and the possibilities they enable, the selective concept is mostly adopted in the last few years. In the selective harvesting of crops, the robotic system firstly decides which are the harvest targets (e.g. ripe fruits identified via a sensory/vision system) and then harvests them. This task may typically involve scanning the whole crop in an orchard or a greenhouse or a part of it, acknowledging and locating targets, cutting/picking them and placing them in a storage unit (e.g. a crate). In the existing literature, there is a variety of integrated robotic solutions that have been proposed towards this end. Representative robotic systems usually consist of: (a) a moving platform/vehicle upon which a robotic manipulator resides, responsible for approaching target fruits/vegetables, grasping and cutting them and subsequently placing them on a crate; (b) a vision system for crop scanning, target identification, detection and localization; (c) a specifically designed robot end-effector to best facilitate target grasping and collection.

\begin{figure}[h!]
    \centering
    \includegraphics[width=.85\textwidth]{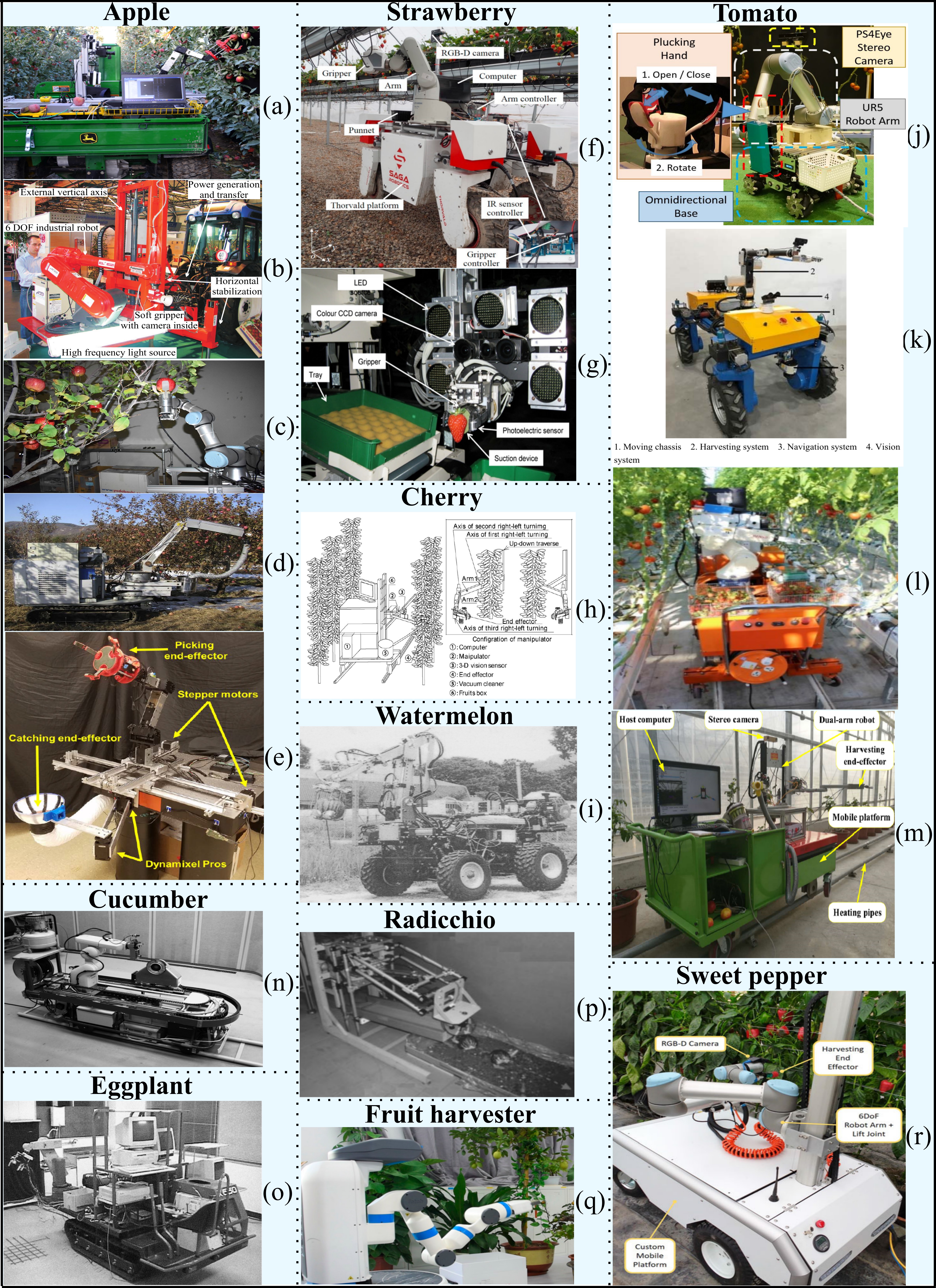}
    \caption{ Robotic harvesting systems;  
    Apple: (a)-\cite{Silwal}, (b)-\cite{Baeten}, (c)-\cite{Onishi}, (d)-\cite{Zhaoda}, (e)-\cite{Davidson}. 
    Strawberry: (f)-\cite{Xiong}, (g)-\cite{Hayashi2}. 
    Cherry: (h)-\cite{Tanigaki}. 
    Watermelon: (i)-\cite{Umeda}. 
    Tomato: (j)-\cite{Yaguchi}, (k)-\cite{Wang}, (l)-\cite{Feng}, (m)-\cite{Ling}. 
    Cucumber: (n)-\cite{VanHenten}. 
    Eggplant: (o)-\cite{sHayashi}. 
    Radicchio: (p)-\cite{Foglia}.
    Fruit harvester: (q)-\cite{tZhang}. 
    Sweet pepper: (r)-\cite{Lehnert1} }
    \label{fig:sys}
\end{figure}

Several harvesting robotic systems for apples may be found in the literature. Since apples have a fairly standard circular shape and a hard nature, they are generally easy to be harvested without significant damage to the fruit. Some of the fastest developed robotic systems report harvest times of 6 seconds \cite{Silwal}-(Fig. \ref{fig:sys}(a)) or 8-10 seconds \cite{Baeten}-(Fig. \ref{fig:sys}(b)) per fruit with a success rate of 80\% or higher. Similar performance may be found in \cite{Bulanon} where harvest is achieved in an average time of 7 seconds with a success rate of 90\%. A slower behavior regarding harvesting time of 15 or 16 seconds may be found in other harvesters, like the fruit harvester in \cite{Onishi}-(Fig. \ref{fig:sys}(c)) that was tested on apples or the apple harvester in \cite{Zhaoda}-(Fig. \ref{fig:sys}(d)) where the importance of real-time obstacle avoidance that the harvesting robots should possess is underlined, due to the obvious complexity and unknown nature of their working environment. 

Furthermore, an approach to potentially reduce the overall cycle time (average time per fruit) of robotic tree fruit harvesting is demonstrated for apple collection by a system employing a pick-and-catch method \cite{Davidson}-(Fig. \ref{fig:sys}(e)). A kinematically redundant picking manipulator with 8 Degrees of Freedom (DoF) performs apple harvesting dropping the apple into the catching end-effector of another 2 DoF robot near the point of fruit detachment \cite{Davidson}. Harvesting tests in a simulated environment with an artificial apple tree located in a laboratory setting showed that the pick-and-catch harvesting method resulted in an over 50\% reduction of average cycle time compared to the pick-and-place method. The system was then briefly tested in a commercial red apple orchard at the end of the 2016 apple harvesting season. Oranges have also been considered as a fruit suitable for automated harvest due to their shape and color and have been a research interest from as early as 1990’s \cite{Muscato}. 

Various robotic solutions may be found in the last decade for the case of strawberries, reporting good performance \cite{Klaoudatos}, \cite{Xiong}-(Fig. \ref{fig:sys}(f)), \cite{Xiong2} and a success rate greater than 75\%  for isolated strawberries with an average harvest time ranging from 6 to 10 seconds per fruit; additional to harvesting operations, such as the packing of the harvested fruits, have also been considered \cite{Hayashi}, \cite{Hayashi2}-(Fig. \ref{fig:sys}(g)). The difference of color between the target fruit and its environment is also exploited in the case of cherries making them relatively easy for detection \cite{Tanigaki}-(Fig. \ref{fig:sys}(h)). In particular, a cherry-harvesting robot was constructed with a 4 DoF arm, integrated with a 3-D vision sensor and a special end-effector that harvested the fruit with its peduncle, to locate the fruits and plan the trajectory of the end-effector avoiding obstacles \cite{Tanigaki}. Researchers have also considered the harvest of heavier fruits like melons \cite{Edan,Bechar2} or watermelons in the case of “STORK” robot \cite{Umeda}-(Fig. \ref{fig:sys}(i)). 

In order to speed up the robot’s harvesting cycle time, multi-arm harvesting operating simultaneously has been explored in \cite{Zion} and in \cite{Ceres}. The work in \cite{Zion} concerns the melon field case, where the arm-melon allocation problem is addressed. A similar approach may be found in the fruit-harvesting robot Agribot \cite{Ceres}, where two robotic arms work in parallel, harvesting the targets that are selected and marked by a human via a point laser range-finder.

Tomatoes are very common and cultivated in a worldwide level, thus one may find several relative works in literature \cite{Yaguchi}-(Fig. \ref{fig:sys}(j)), \cite{Wang}-(Fig. \ref{fig:sys}(k)), \cite{Feng}-(Fig. \ref{fig:sys}(l)). The dual-arm (bi-manual) robot harvester \cite{Zhao}, \cite{Ling}-(Fig. \ref{fig:sys}(m)) stands out in which one manipulator cuts the fruit and the other picks it up, contrary to most other agricultural robotic systems that utilize only one arm (uni-manual).

The importance of developing and adapting new high productivity cultivation systems is discussed in the literature, modifying the crop and its envelopment to better fit for robotic harvesting. In particular, the high-wire system is considered for the cucumber case \cite{VanHenten}. This issue is also raised and investigated in \cite{vanHerck} in the SWEEPER\footnote{\href{https://cordis.europa.eu/project/id/644313}{https://cordis.europa.eu/project/id/644313} \& \href{http://www.sweeper-robot.eu/}{http://www.sweeper-robot.eu/} - Last accessed: 22-11-2022} project. Such specifically designed cultivation systems can facilitate even further the automation in the agriculture, making robotic harvesting easier and faster \cite{Bloch}, e.g. the V-trellis fruiting wall architecture for apples \cite{Silwal}, enhancing target fruit access and presenting fewer interfering obstacles.

In addition to fruits, vegetables have also been considered as harvest targets by automated harvesting machines. An autonomous robot for cucumber harvesting is proposed in \cite{VanHenten}-(Fig. \ref{fig:sys}(n)), reporting a success harvesting rate of 80\% and a cycle time of 45 seconds for each harvest target. In the eggplant case \cite{sHayashi}-(Fig. \ref{fig:sys}(o)), 62.5\% success rate and an average harvest time of 64.1 seconds are reported. Works regarding radicchio \cite{Foglia}-(Fig. \ref{fig:sys}(p)) and asparagus \cite{Irie} may also been found in the literature. Relatively low success rates in crops detachment are reported in \cite{tZhang}-(Fig. \ref{fig:sys}(q)), where the presented autonomous harvester for crops with peduncles (either vegetables or fruits) is tested on both plastic and real crop cases, exhibiting a 67\% and a 52\% success rate respectively. A research interest has been shown for sweet peppers as well, in \cite{Kitamura}, \cite{Lehnert1}-(Fig. \ref{fig:sys}(r)) with the “Harvey” robot and in \cite{Arad}. 
Specifically, \cite{Arad} is part of the EU project SWEEPER that targeted towards the development and testing of a practical robotic harvesting solution for sweet peppers in real-world conditions, reporting an average harvest time of 24 seconds per pepper. 

Summarizing, as can be seen in Table \ref{table:sys}, it becomes clear that the majority of harvesting robotic systems involve crops with shape, size, rigidity and color that facilitates automated harvesting. The lack of any reported harvesting solutions for grapes which is a high-value crop indicate the existence of specific challenges in this case. Nevertheless, various research works may be found for the case of grapevines. 

As part of the PIC4SeR\footnote{Politecnico Interdepartmental Centre for Service Robotics - Politecnico di Torino, \href{https://pic4ser.polito.it/}{https://pic4ser.polito.it/} – Last accessed: 22-11-2022} mission for intelligent vineyard healthcare, the Agri.q UGV is developed in \cite{Quaglia,Quaglia2,Quaglia3}, with its operation extended in other types of orchards as well. Overcoming terrain irregularities or slopes and providing various services such as crop health monitoring, collection of soil and/or plant samples or application of fertilizers, this robotic solution investigates the innovative possibility of multiple robotic systems working collaboratively for coordinated field monitoring and servicing, including a UGV and a UAV (drone). An early attempt for developing a multipurpose agricultural robot working in vineyards can be found in \cite{Monta}, considering various tasks such as berry thinning, spraying, bagging and harvesting. The robotic manipulator developed in \cite{Monta} was also utilized in \cite{Ogawa}, where the explicit task of spraying chemicals under grapevine trellis for crop health/disease removal purposes is addressed. Disease detection and selective spraying for production quality enhancement is the research target of \cite{Oberti} as well, part of the EU-project CROPS\footnote{\href{https://cordis.europa.eu/project/id/246252}{https://cordis.europa.eu/project/id/246252} \& \href{http://www.crops-robots.eu/}{http://www.crops-robots.eu/} - Last accessed: 22-11-2022} whose main goal was to develop, optimize and demonstrate a multipurpose agricultural robotic system \cite{Best}. Project CROPS attempted to achieve grape harvesting as well, however with no reported success. A similar application, i.e. vineyard protection, was also targeted by the GRAPE\footnote{\href{https://echord.eu/grape/}{https://echord.eu/grape/} \& \href{https://www.grape-project.eu}{https://www.grape-project.eu} - Last accessed: 22-11-2022} project. In GRAPE, an autonomous ground robot \cite{Roure} with a robotic arm was developed for health monitoring and automatic pheromone dispenser distribution for plague control. Other works in literature involve tasks like weeding i.e. removal of weeds in areas of a vineyard \cite{Reiser}, monitoring i.e. localization and mapping of vineyard for crop monitoring \cite{dosSantos} and pruning i.e. cutting off some of the grapevine older canes/branches while leaving others that are healthier \cite{Botterill}. Through utilization of two independent robots working cooperatively (harvesting and inspection robots), ongoing EU funded project BACCHUS\footnote{\href{https://cordis.europa.eu/project/id/871704}{https://cordis.europa.eu/project/id/871704} \& \href{https://bacchus-project.eu/}{https://bacchus-project.eu/} - Last accessed: 22-11-2022} aims at contributing in the limited field of automated grape harvesting. Introducing a human-like harvesting approach via a bi-manual setup (harvesting robot) contrary to most existing solutions, while combining it with crop inspection, mapping, as well as vital data collection e.g. concerning grape health/maturity (inspection robot), BACCHUS targets on pushing forward the application of precision agriculture in a particularly challenging environment such as a vineyard (BACCHUS as well as the previously mentioned GRAPE, CROPS and SWEEPER will be described further in Section \ref{ch4}, where the research projects are presented in details).
\begin{table}[h!]
\centering
\begin{tabular}{ |p{1.5cm}||p{3.5cm}|p{5.5cm}|  }                                             \hline
Crop & Citation & Characteristics                                                          \\ \hline
                                                                                              \hline 
\multicolumn{3}{|c|}{\textbf{Harvesting}}                                                  \\ \hline

\multirow{4}{1.5cm}{ Apple }                                                                       & 
\multirow{4}{3.5cm}{ \cite{Silwal, Baeten, Bulanon, Onishi, Zhaoda, Davidson} }                    & 
\multirow{4}{5.5cm}{Harvest times / success rates: 6-16 sec / 80-90\%. Methods for faster/easier harvest: e.g. pick-and-catch harvesting \cite{Davidson}. V-trellis farming architecture \cite{Silwal}.}  
                                                                      \\ & & \\ & & \\ & & \\ \hline

\multirow{2}{1.5cm}{Orange}                                                                        & 
\multirow{2}{3.5cm}{\cite{Muscato}}                                                                & 
\multirow{2}{5.5cm}{Early interest shown (1990's). Harvest-friendly crop shape and color.}                                                                                                      \\ & & \\ \hline

\multirow{2}{1.5cm}{Strawberry}                                                                    & 
\multirow{2}{3.5cm}{\cite{Klaoudatos, Xiong, Xiong2, Hayashi, Hayashi2}}                           & 
\multirow{2}{5.5cm}{Success rates $>$ 75\%. Harvest times: 6 to 10 sec. Packing operation \cite{Hayashi,Hayashi2}.}                                                                                                \\ & & \\ \hline

\multirow{2}{1.5cm}{Cherry}                                                                        & 
\multirow{2}{3.5cm}{\cite{Tanigaki}}                                                               & 
\multirow{2}{5.5cm}{Different fruit/environment color exploited to facilitate detection.}                                                                                                                          \\ & & \\ \hline

\multirow{2}{1.5cm}{Melon}                                                                         & 
\multirow{2}{3.5cm}{\cite{Edan, Bechar2, Zion}}                                                    & 
\multirow{2}{5.5cm}{Simultaneous multi-arm harvesting \cite{Zion} to speed up harvest.}                                                                                                                            \\ & & \\ \hline

\multirow{2}{1.5cm}{Watermelon}                                                                    & 
\multirow{2}{3.5cm}{\cite{Umeda}}                                                                  & 
\multirow{2}{5.5cm}{Reported success rate: 66.7\%; robot weighting $<$ 300 kg picking fruits of 13 kg.}                                                                                                            \\ & & \\ \hline

\multirow{2}{1.5cm}{Tomato}                                                                        & 
\multirow{2}{3.5cm}{\cite{Yaguchi, Wang, Feng, Zhao, Ling}}                                        & 
\multirow{2}{5.5cm}{Dual-arm harvester \cite{Zhao,Ling} contrary to common uni-manual systems.}  
                                                                                    \\ & & \\ \hline

\multirow{3}{1.5cm}{Cucumber}                                                                      & 
\multirow{3}{3.5cm}{\cite{VanHenten}}                                                              & 
\multirow{3}{5.5cm}{Success rate: 80\%; cycle time: 45 sec. High-wire cultivation system for easier robotic harvesting.}                                                         \\ & & \\ & & \\ \hline

\multirow{2}{1.5cm}{Eggplant}                                                                      & 
\multirow{2}{3.5cm}{\cite{sHayashi}}                                                               & 
\multirow{2}{5.5cm}{62.5\% success rate; 64.1 sec average harvest time.}            \\ & & \\ \hline

\multirow{1}{1.5cm}{Radicchio}                                                                     & 
\multirow{1}{3.5cm}{\cite{Foglia}}                                                                 & 
\multirow{1}{5.5cm}{Average cycle time below 7 sec.}                                       \\ \hline

\multirow{1}{1.5cm}{Asparagus}                                                                     & 
\multirow{1}{3.5cm}{\cite{Irie}}                                                                   & 
\multirow{1}{5.5cm}{Average harvesting time: 12 sec.}                                      \\ \hline

\multirow{2}{1.5cm}{Sweet pepper}                                                                  & 
\multirow{2}{3.5cm}{\cite{Kitamura, Lehnert1, Arad}}                                               & 
\multirow{2}{5.5cm}{\cite{Arad} part of SWEEPER project; average harvest time: 24 sec.}                                                                                                                            \\ & & \\ \hline

\multicolumn{3}{|c|}{\textbf{Monitoring/Vineyard Management}}                              \\ \hline
 
\multirow{6}{1.5cm}{ Grape }                                                                       & 
\multirow{6}{3.5cm}{ \cite{Quaglia, Quaglia2, Quaglia3, Monta, Ogawa, Oberti, Best, Roure, Reiser, dosSantos, Botterill} }                                                                            & 
\multirow{6}{5.5cm}{Health monitoring (e.g. \cite{Roure} of GRAPE project), yield estimation, soil/plant sampling, fertilizer application, selective spraying (e.g. \cite{Oberti} of CROPS project), thinning, weeding, pruning, vineyard mapping.}                          
                                                        \\ & & \\ & & \\ & & \\ & & \\ & & \\ \hline
\end{tabular}
\caption{Summary of harvesting and other agricultural systems}
\label{table:sys}
\end{table}

%-----------------------------------------------------------------------------------------------------------------------------------------------------

\section{Robotic Functionalities and Hardware for Automated Harvesting}  \label{ch3}

\begin{figure}[h!]
    \centering
    \includegraphics[width=.75\textwidth]{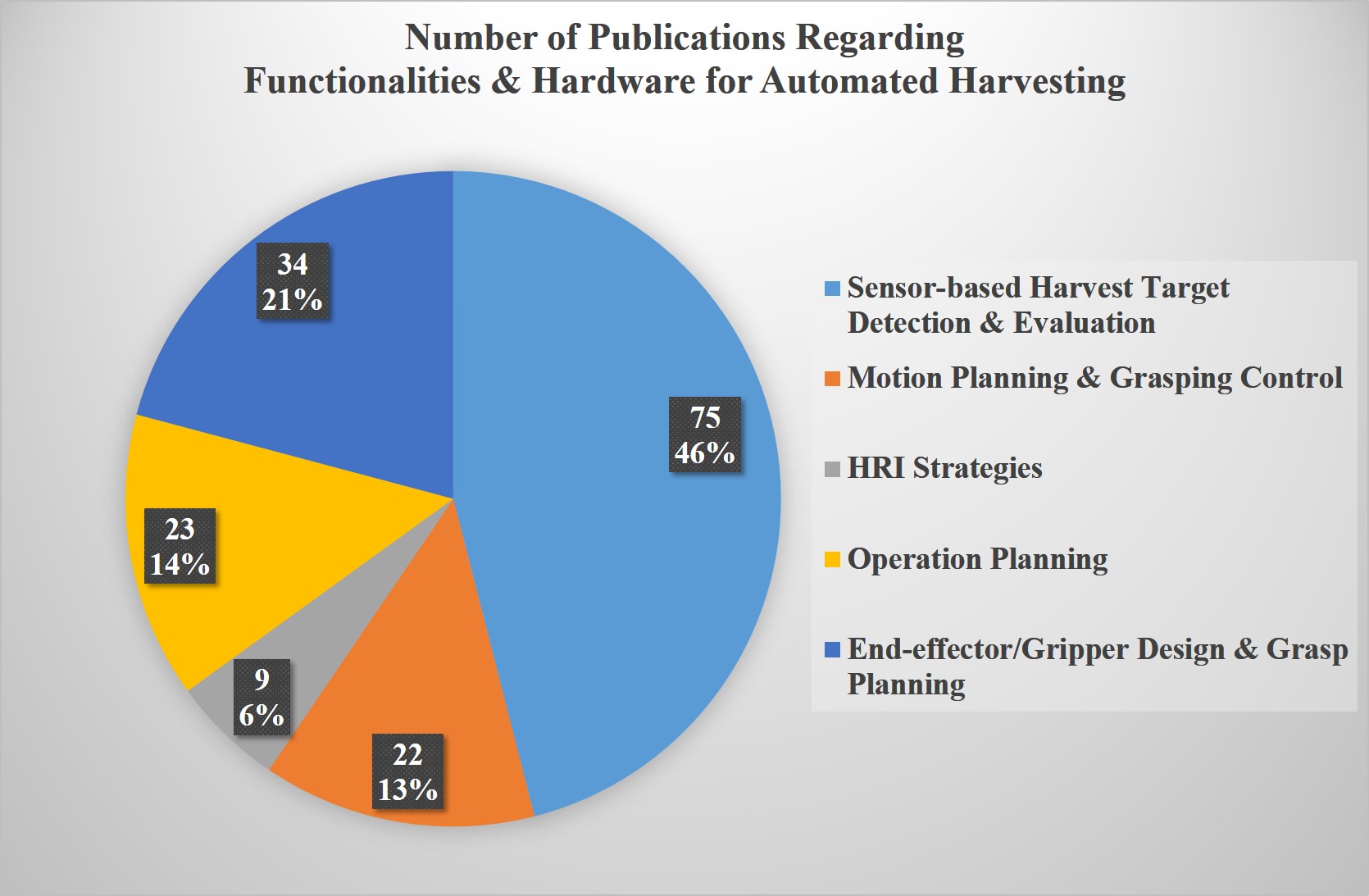}
    \caption{Distribution of functionality-related works in literature}
    \label{fig:func_no}
\end{figure}

Apart from the development of a complete robotic system for the automated harvesting of fruits or vegetables, one may find a significant amount of works in the literature that deal with solutions over specific challenges related to the overall harvesting task. Vision-related challenges regarding harvest targets detection and their maturity and health evaluation, navigation of robotic systems in fields and orchards, motion and grasp planning of the robotic system’s manipulator for successful harvesting, design of end-effectors with specific requirements depending on the harvest crops as well as operation planning and human robot interaction (HRI) systems for the agricultural domain are some of the problems addressed in the literature. Fig. \ref{fig:func_no} depicts how the referenced  publications are distributed with respect to their targeted functionality, indicating that the majority of them address crop detection and evaluation. 

%###################################################################################################
\subsection{Sensor-based Detection and Evaluation of Harvest Targets}  \label{subch31}
%###################################################################################################

The detection of the harvest targets as well as the crops’ health and maturity evaluation are of high importance for the farmers concerning both estimating the yield quantity and quality, with evident impacts on the financial earnings and managerial issues (e.g. arrangements for crop shipment or storage), and the actual application of robotic harvesting methods. To this end, robotic systems with various visual sensors/cameras (e.g. binocular vision, laser vision, kinect camera, multispectral camera etc.) placed on the robotic platforms themselves and/or mounted on the robotic manipulator’s end-effector (eye-in-hand) are proposed, while image processing methodologies are employed to transform the raw imagery data to valuable knowledge \cite{Tang}. The main tasks addressed in the crop-detection literature refer to the fruit recognition, which is the key issue in the vision-guided autonomous fruit harvesting and yield estimation systems, the crop’s maturity/health evaluation, which is critical on the crop’s suitability for harvesting decision-making, and the environment modelling and surveying, which provides useful input for the autonomous agricultural applications.

%###################################################################################################
%###################################################################################################
\subsubsection{Vision-based Crop Detection} \label{subch311}
%###################################################################################################
%###################################################################################################

Over the last years, a great variety of research work has been conducted regarding vision-based detection methods for utilization in various automated agriculture applications \cite{saleem}. Adaptive thresholding algorithms \cite{Vitzrabin}, methods based on texture, color and shape cues \cite{Blokp,Kurtserp,Nuske}, combination of RGB and depth information \cite{Luo}, reinforcement-learning approaches \cite{Ostovar} and deep convolutional neural networks \cite{Altaheri,Ge} are used for image segmentation into background and foreground in order to detect target fruits. Additionally, to properly position the robot with respect to the target, localization of the harvest targets may also include estimation of their orientation \cite{Lehnert2,Kang,Zapotezny} leading to the robotic end-effector approaching its target in a way that is better suited for grasping/cutting the peduncle.

Harvest targeted fruit detection is a task that faces multiple challenges due the dynamic and unstructured nature of both the crops and the outdoors environment. Therefore, to cope with color and texture-based sensitivities to the variety of illumination conditions, deep learning approaches \cite{Arad2} and RGB-D data methods are proposed \cite{Tian2,Kurtser}. Moreover, occlusions that often occur due to leaves of branches affect directly the harvest-target detection success. To address this problem image analysis including per-pixel segmentation and region detection is used in \cite{McCoolc}, while a multi-class approach that splits citrus samples into different cases (normal, branch occlusion, leaf occlusion, slight occlusion, overlapping and main brunch) is applied in \cite{Liu}. Furthermore, the authors of \cite{Lehnert} propose an active vision approach that utilizes 9 eye-in-hand RGB cameras to solve the “next best view problem” for sweet peppers occluded by leaves, while \cite{Xiong2} combines vision with specific movements that the robotic manipulator’s end-effector performs in order to push away any surrounding obstacles that occlude the harvest target. Apart from the difficulties occurring from the environment, there are challenges created by the fruit’s nature. In this scope, the challenging color of certain fruits, such as green grapes, is handled using different color spaces depending on the kind of crops and considering additional features during the image process analysis for the detection/localization task (\cite{Zemmour,Vitzrabin2}, parts of SWEEPER and CROPS projects respectively). Moreover, vision issues such as the lack of contrasting features on smooth round-surfaced fruits that otherwise could help in their detection is dealt by utilization of the gradual variation of intensity and the gradient orientation upon the fruit’s surface \cite{Pothen}.

A group of works on agricultural object detection utilizes deep neural architectures for semantically segmenting the scene in conjunction with methods for clustering the resulting segmentation map. The authors of \cite{Bargoti} compute the semantic segmentation of apple orchard RGB images using multi-scale Multi-Layered Perceptrons (MLPs) and Convolutional Neural Networks (CNNs) that are extended by incorporating metadata, such as sun position, while on top of the segmentation the Watershed and the Circular Hough Transform algorithms are utilized to perform fruit detection. The authors of \cite{Lin} detect guava using a Fully Convolutional Network for semantically segmenting RGB images and then extract clusters that correspond to individual fruits. In \cite{Fawakherji}, Unet is utilized to semantically segment RGB images into soil/vegetation classes, then blob extraction is applied, and finally the results are classified into weed/crop using a CNN classifier with VGG16 backbone. Moreover, in \cite{Hani} a Unet followed by connected component analysis, a Faster-RCNN with FPN and a focal loss term, and a Gaussian Mixture Model are compared for fruit detection in apple orchards in order to achieve 3D reconstruction of the trees. A fruit-counting algorithm is presented in \cite{Liu1} that incorporates an object detection step with FCN based semantic segmentation and a contour detection step. Furthermore, the authors of \cite{gLin} utilize a probabilistic image segmentation method to segment RGB fruit images in foreground/background, mask the corresponding depth image with the probabilistic segmentation output, and then employ a region growing method to obtain clusters from the depth data and detect fruits through an SVM classifier. Some works focus solely on the problem of semantically segmenting the captured agricultural scene. In \cite{McCool}, the authors employ a sliding-window Inception-v3 classifier to segment RGB images in three classes: crop, weed, and background. 

Many authors utilize well-known one-stage or two-stage deep CNN detectors, using the algorithms either off-the-shelf or with custom modifications, to directly detect fruits and/or plants in their input data. In this scope, the authors of \cite{Santos} compare YOLOv2, YOLOv3 and Mask-RCNN in order to implement a fruit counting graph-based application, while various applications of YOLOv2 and YOLOv3 are also used in \cite{Koirala,Tian} for mango and apple detection. The authors of \cite{Gonzalez} use Mask-RCNN for blueberry detection and quantification, investigating different feature extractors/backbones for the task, while in \cite{Ganesh}, a Mask-RCNN with RGB \& HSV 6-channel input is utilized to detect oranges. Moreover, Faster-RCNN is used in \cite{Sa,GeneMola,Hasan} for the detection of multiple types of fruits based on RGB and Near-Infrared images, the identification of Fuji apples on RGB-D images and the detection of 4 different wheat growth stages respectively. In \cite{Kang2}, an automatic label generation module named Clustering-RCNN, along with the one-stage deep learning object detector LedNet are developed for the detection of apples in RGB images of orchards, while the same authors propose in \cite{Kang3} a DaSNet, which utilizes a Gated Feature Pyramid Network for multi-level feature fusion, ASPP for multi-scale information fusion, and a lightweight designed backbone. A strawberry detector is employed in \cite{Kirk} based on a modified RetinaNet network that utilizes ResNet-18 as its backbone, FPN for multi-level feature fusion and takes as input a concatenated 6-channel RGB + CIELab representation of the image, while in \cite{Dias} a novel object detector for flowers is proposed, based on regions generated by the SLIC algorithm that are used from a CNN architecture to extract features that are fed into a SVM in order to detect apple flowers.

%###################################################################################################
%###################################################################################################
\subsubsection{Field/Orchard Modelling and Surveying} \label{subch312}
%###################################################################################################
%###################################################################################################

Besides the aforementioned work that focuses on the crops, some researchers investigated issues related to the whole agricultural environment, i.e. the fields and orchards. The authors aim at modelling or surveying the environment using various visual sensors (e.g. spectral cameras, RGB, RGB-D, LiDAR and thermal imaging sensors) while it can be observed that phenotyping has attracted the interest of many authors. Phenotyping refers to the quantitative description of a plant’s properties and aims at the holistic evaluation and characterization of its performance. To this end, a mobile platform for monitoring Canola plants is presented in \cite{Bayati}. The platform carries different sensors, cameras and other measurement equipment while it geo-tags the acquired data using GPS for future retrieval. Moreover, a fully automated robotic platform moving on rails and mounted with various sensors (chlorophyll fluorescence and thermal infrared cameras, two hyperspectral imagers and dual 3D laser scanners) is established in \cite{Virlet} in order to facilitate continuous and high-throughput monitoring of crop performance. An autonomous ground vehicle named “Vinobot” and a mobile observation tower named “Vinoculer” are introduced in \cite{Shafiekhani} for the inspection of large field areas during day and night in order to provide high-throughput plant phenotyping. The ground vehicle collects data from individual plants, while the observation tower oversees an entire field, identifying specific plants for further inspection by the Vinobot. Furthermore, the authors of \cite{sGutierrez} address the grapevine varietal classification while avoiding the destructive techniques of classic ampelography. Particularly, an on-the-go hyperspectral imaging system is proposed that, by means of SVM and MLPs, successfully classifies leaves from different grapevine varieties.

To achieve field and orchard monitoring and analysis, image processing methods are commonly used. Therefore, semi-supervised weed classification based on CNNs is proposed in \cite{Hu}, while an anomaly detection system is presented in \cite{Christiansen} that uses the features derived from a CNN classifier to detect anomalies in agricultural fields. Moreover, the authors of \cite{Isokane} utilize multi-view images and a probabilistic setting to infer the existence of branch structures hidden under leaves and to reconstruct them using particle flow simulation, whereas in \cite{Botterill}, the 3D model of the whole orchard is reconstructed by means of stereo cameras.  A system able to recognize the canopy volume is proposed in \cite{Bietresato} using two LiDAR sensors aligned vertically for scanning the same targets so as to provide information for the pruning task. Furthermore, in \cite{Williams}, an automated method performed on hyperspectral images is developed in order to segment raspberry plants from the background using a selected spectral ratio combined with edge detection while graph theory is used to detect the continuous boundary between uninteresting plants and the area of interest.

%###################################################################################################
%###################################################################################################
\subsubsection{Crop Evaluation} \label{subch313}
%###################################################################################################
%###################################################################################################

A vital issue during the harvesting season is the evaluation of the fruits mainly in terms of maturity estimation. In this scope, several works have explored the use of hyperspectral cameras so as to achieve non-destructive evaluation approaches. Therefore, VIS-NIR spectroscopy has been mainly used to examine point spectroscopy in laboratory environments. In \cite{andrj,dosSantos2,Porep,Giovenzana}, high-resolution spectrometers capture single spectral signatures from grape berries and bunches under specific illumination conditions to achieve maturity evaluation. On the other hand, there are also efforts to use hyperspectral cameras in orchards while applying methods that handle the shadows. In particular, the authors of \cite{Wendel} use a ground vehicle called “Shrimp” equipped with a hyperspectral camera to estimate the dry matter content (DM) of mangos while LiDAR data are used for tree segmentation so that DM predictions are projected to individual trees. The authors extend their work in \cite{Gutierrez2, Gutierrez3} with pre-processing techniques for illumination compensation while they explore different optical filters for mango DM estimation.

Apart from the fruit maturity, authors have explored the evaluation of fruits regarding other criteria, such as checking for damages. In this scope, a modified version of YOLOv3 is used in \cite{Sa} that incorporates DenseNet layers to detect apple lesions in a dataset that has been augmented by synthetic images created from CycleGAN, while in \cite{Nasirahmadi} YOLOv4 CSPDarknet53 method is proposed for detecting damages in sugar beet. Using VIS/NIR transmission coupled with diameter correction and a deep 1D-CNN, the early detection of freezing damage in oranges is achieved in \cite{Shiijie}. The authors of \cite{Jin} implement a per-pixel classification/semantic segmentation of hyperspectral healthy and diseased wheat data, experimenting with two different input configurations in a Deep Convolutional Recurrent Neural Network. Furthermore, the authors of \cite{Mack} provide a complete and detailed crop 3D reconstruction through image data analysis to be used for phenotyping in order to further explore the crop’s features such as disease resistance, breeding efficiency etc.

%###################################################################################################
%###################################################################################################
\subsubsection{Detection and Evaluation of Harvest Targets Using Drones} \label{subsubch314}
%###################################################################################################
%###################################################################################################

Crop detection and evaluation functionalities by drones have shown a wide potential in precision farming \cite{Hafeez_drone} which, supplementary to ground robots could further boost automation in the farming domain. Mainly focusing on crop monitoring and weed/disease detection, the drone applications mostly use three types of sensors: 1) RGB, 2) multispectral and 3) hyperspectral sensors \cite{Esposito_drone}. Based on RGB input, the authors of \cite{Su_drone} monitor fields for yellow rust using U-Net deep learning networks while CNNs are used in \cite{Crimaldi_drone} for detection of plant diseases that affect their leaves. Moreover, weed detection takes place using a Bag –of Visual –Words framework in \cite{Pflanz_drone} and various segmentation methods (CGNet, ENet, ERFNet) are compared in \cite{Jiacai_drone} towards dandelion identification. Using multispectral sensors, weed detection and crop monitoring is also explored in \cite{SaI_drone} and \cite{Mazzia_drone} while, due to the availability of the high number of radiometric bands, \cite{Candiago_drone} extends to the extraction of non-visible field features such as early stage plant disease and soil water content. Furthermore, hyperspectral data is used in \cite{Ya_drone} to identify the most significant spectral bands and perform weed classification based on discriminant analysis as well as in \cite{Mink_drone} to evaluate maize and weed response to herbicide applications.

%###################################################################################################
\subsection{Motion Planning and Control for Grasping and Picking a Harvest Target}  \label{subch32}
%###################################################################################################

Motion planning and control of a harvesting robotic system has been addressed by various works in
literature regarding either guiding the system’s vehicle/platform inside the orchard (between
plants) or planning of the system’s robotic manipulator motion in order to approach and pick a harvest
target. When guiding a robotic vehicle/platform in such a difficult environment as an orchard with many
obstacles (e.g. plants, trees, branches), utilization of feedback provided by various sensors is essential.
Global Positioning System (GPS), accelerometers and gyroscopes (Inertial Navigation System – INS),
vision i.e. cameras, visual odometry (VO), as well as a fusion of the above feedback data \cite{Zaidner} are often considered. 

Several works address the problem of robot’s navigation within fields or orchards. The development of a navigation system that can guide a field robot to travel from a farm station to a citrus grove and visit each tree autonomously with obstacle avoidance ability is the subject of \cite{Gan}. GPS, IMU and the wheel encoders are integrated by an extended Kalman filter to provide accurate odometry information. Then, the SLAM algorithm combines the odometry data with the LiDAR data, enabling the robot to do path planning and navigating based on laser scans. To facilitate navigation, the robotic platform “BoniRob” \cite{Biber} for crop scouting, developed by AMAZONE\footnote{\href{http://www.amazone.de}{http://www.amazone.de} - Last accessed: 22-11-2022}, utilizes the FX6 laser to detect the ground from a 3D point cloud recorded by FX6. Furthermore, convolutional neural networks in combination with visual-based feedback from a camera are also proposed for crop identification, driving path planning and fitting between plant rows \cite{yGu}.
 
Concerning map generation, there are many works for indoor environments which give satisfactory results \cite{Ouellette}. Nevertheless, there are still few proposals concerning the mapping in agricultural environments. Their goal is to decrease navigation errors \cite{jmZhang} as well as to enable the vehicle to return to specific locations and perform tasks, such as spraying, in a suitable and precise manner, thus saving valuable resources \cite{Libby}. Such example is described in \cite{jtJin} where a mapping strategy based on maize plants detection using a stereo camera is presented. In recent years, many works have been made to solve the problem of simultaneous localization and mapping (SLAM), providing an appealing alternative to user-built maps and showing that robot operation is possible in the absence of an ad hoc localization infrastructure \cite{Cadena}. Cheein et al. \cite{Auat} developed a precision SLAM algorithm based on extended information filter (EIF-SLAM) for agricultural environments, in their case olive groves. The most significant features in the agricultural environment are used for the optimization and pose estimation process. Moreover, the use of a SLAM algorithm, called graph-SLAM, is proposed in \cite{Pierzchala} as a means to generate cost-effective local maps of forests. In particular, the 3D map is generated from laser scans mounted on a mobile platform, first by relying on laser odometry and then by improving it with robust graph optimisation after loop closures.

Handling the manipulator’s motion towards the harvest target is considered a rather challenging task taking into consideration the highly unstructured environment of orchards. CROPS project \cite{tNguyen} is a representative work that deals with most of the typical issues arising during the manipulator’s motion towards the harvest target by approaching it dynamically taking into consideration possible obstacles. Collision maps are typically generated and can be also updated in real-time as the robot moves, often including 3D point cloud data \cite{Hornung} provided by feedback from the vision system and other sensors the overall robotic system may be equipped with. These maps are utilized so that a collision free trajectory is planned for each of the robotic manipulator’s joints. Guiding the robot end-effector utilizing information provided by eye-in-hand sensing as in the cases of \cite{Mehta} and \cite{Barth,Ringdahl} (parts of CROPS and SWEEPER projects respectively) is indeed a very common approach for visual servoing. In such approaches, information concerning the best approach direction to maximize target’s detection rate may be highly valuable \cite{Ringdahl2}. It is evident that visual servoing can be sensitive in camera calibration parameter uncertainties, as illustrated in \cite{Camacho} where the combined utilization of image-based visual servoing (IBVS) for the harvest target approach phase and position-based visual servoing (PBVS) for making fine adjustments in the robot’s movement is proposed to cope with such uncertainties.

After the harvest target is reached and grasped by the robot’s end-effector, its picking may be achieved via a detaching motion that the robot performs in order for the target fruit to be released from the tree’s branch. This motion may vary with the obvious goal to detach the harvest target as easily as possible without causing any damage to it. Four basic picking patterns including horizontal pull, vertical tension, bending and twisting are performed in \cite{Bu} in order to analyze the effects of different picking patterns on fruit detachment. The tension parameter, which includes the horizontal pull and vertical tension, is found to be the dominant factor during the detachment process, with the vertical pull possibly leading to the pull-out of the stem. Optimized results indicate the horizontal pull with a bending and twisting motion as the potential optimum combination. Picking patterns for successful crop detachment are directly connected to the physical properties of harvest targets. To this end, works may be found in literature that focus their research to target crops themselves, investigating their rheological characteristics (deformation) \cite{jWei} and physical properties regarding e.g. how much force may be exerted upon them without damaging or bruising them. In \cite{jWei}, apples are considered with the ultimate purpose of designing an impedance controller for the robotic system of \cite{Zhaoda} to compliantly pick them, while in \cite{Roshanianfard} pumpkins are investigated in order to find an appropriate harvesting procedure that will not damage them. A major challenge in robot motion planning that one may easily consider is the possible harvest target motion (e.g. a fruit may move due to gusts of wind that may shake the tree’s branches). In \cite{ssMehta}, such a problem is addressed by proposing a robust, image-based, non-linear visual servo controller (along with its Lyapunov stability analysis) that regulates the end-effector to the target fruit’s location, compensating for unknown disturbances like the fruit’s motion. 

%###################################################################################################
\subsection{HRI in Agriculture}  \label{subch33}
%###################################################################################################

Agricultural processes are mostly performed by human-operated machines while few are handled by autonomous robots. However, there are applications that are difficult to be fully automated while the efforts in this direction appear to be on a long-term path. In this scope, HRI strategies in agriculture have attracted the scientific interest in order to address such complex problems, providing secure, faster and more productive solutions than the traditional approaches \cite{Vasconez}.

Several works in agricultural HRI literature focus on activities such as facilitating farming robots’ navigation in fields and orchards, detection of fruits and vegetables, spraying \cite{Bechar3}. In particular, different user interface modes for target recognition and spraying are explored in \cite{Adamides}. The robot navigation along vineyards can be handled by the user when needed, views from the on board camera sensors are visualized while teleoperation is achieved through three different interface configurations which use a mouse, a Wii remote\footnote{\href{https://www.nintendo.co.uk/Support/Wii/Usage/Wii-Remote/Basic-Operations/Basic-Operations-243993.html}{https://www.nintendo.co.uk/Support/Wii/Usage/Wii-Remote/Basic-Operations/Basic-Operations-243993.html} - Last accessed: 22-11-2022 } 
and a digital pen. Moreover, a web based UI for controlling autonomous vehicles for apple harvesting is proposed in \cite{Bergerman}. The user can select the field area of interest, the vehicle’s speed, and decide whether certain orchard’s rows should be skipped or traversed multiple times. The benefits of the use of HRI strategies aided with visualization technologies for melon detection are analysed in \cite{Bechar2}. Specifically, detection increased by $4$\% in comparison to manual detection and by $14$\% compared with a fully autonomous approach, reaching average detection rates between $94$\% and $100$\%. Moreover, the detection times were $20$\% shorter than the ones achieved manually.

Designing complex applications for HRI in agriculture requires deep understanding of the users’ cognitive factors and the task demands. In this scope, using a knowledge engineering approach, the authors of \cite{Cullen} proposed a work model based on information obtained from operators to evaluate the mowing task in a citrus grove. Moreover, a method to estimate the operator’s mental workload in multiple information presentation environment for agricultural vehicles is presented in \cite{Jin17}. The aim of the method is to facilitate the design and optimize the human-machine interface resources to ensure highly-efficient and safe operations in the fields. Alternately, other works use sensors to estimate the human operator’s cognitive status. Such example is provided in \cite{GomezGil} where non-invasive sensors, such as EMG (electromyography), are used to study and analyse the tractor handling performance in agricultural fields while in \cite{Szczepaniak} the tractor operator’s driving characteristics are modeled through simulation so as to achieve better adaptation of the vehicle’s dynamic properties to the driver.

%###################################################################################################
\subsection{Operation Planning for Crop Inspection and Harvesting}  \label{subch34}
%###################################################################################################

Production scheduling is a critical part of production management that has preoccupied the scientific community over the last decades leading to prolific literature. The researchers have investigated both static and dynamic scheduling scenarios \cite{Mohan} while the studied cases involve job-shops, multi-agent systems, robot-human cooperation, single robots operating multiple tasks \cite{Petrovic} etc. Over the years, various approaches \cite{Turkyilmaz} have been used to deal with the optimization of task planning including particle swarm operations \cite{Xu}, artificial bee colonies \cite{Li}, variable neighborhood search \cite{Zheng}, Tabu search \cite{Li2}, ant colony optimization \cite{Huang}, evolutionary algorithms \cite{Reddy,Huang2} and other heuristic approaches \cite{Ojstersek,Zhou}.

In the robotic agricultural domain, the operations planning applications include planning of harvesting actions, team coordination of autonomous vehicles, optimization of field area coverage and route planning. Moreover, human-robot cooperation is also investigated, nevertheless, robots act, in these cases, as crop transporting vehicles scheduled to provide optimal support to human harvesters \cite{Seyyedhasani}. In particular, a stochastic simulation based on a finite state machine is used to evaluate human robot collaboration and provide to the strawberry-harvesting workers the optimal support with automated transport of the collected fruits. Regarding the optimal route/area covering planning, a general approach for fleets of autonomous vehicles is suggested in \cite{Conesa}. Taking into account different criteria, such as vehicles with different characteristics, field variabilities and the refilling possibility, the authors deal with this combinatorial optimization problem using the metaheuristic search of simulated annealing. Furthermore, in \cite{Edwards} multiple machines are scheduled to work on various operations in multiple areas based on the fields’ readiness using Tabu and modified Tabu search technique. In \cite{Ahsan}, a mobile seed refilling system that feeds multiple planters during large-scale field operations optimizes its route using a stochastic solution based on a genetic algorithm. Moreover, the travelling salesman problem methodology is applied on \cite{Jensen,Santoro} and \cite{Cheein} for the optimization of coverage planning for capacitated fields operations and route planning for sugarcane and avocado harvesting respectively. In \cite{Richards15}, the field area coverage is handled by the users themselves through a UI that allows them to schedule the robots' routes taking also into account refilling stations to replenish their resources (e.g. pesticide and energy).

Apart from the applications focusing on the optimization of area/route coverage according to the fields’ properties and the available means, several researchers explored the optimization of the actual harvesting actions. In this case, the robot is parked in front of the plant of interest and has to collect the available fruits with one or multiple arms. In \cite{Mann}, a multi-arm robotic melon harvester optimized its performance utilizing an extension of graph coloring. In particular, the robot scheduled the fruit collection and decreased the harvest time by selecting the most efficient actuators, number of manipulators and robot’s velocity. Similarly, in \cite{Barnett}, multiple arms of a kiwi robot collector are assigned with tasks based on a cluster approach. In \cite{Kurtser2}, one manipulator is utilized to perform sensing and harvesting tasks simultaneously. The sequence of these actions is scheduled through the travelling salesman paradigm taking into account the actions’ costs and the traveling time.

%###################################################################################################
\subsection{End-Effectors Development, Automated Gripper Design and Grasp Planning for Harvesting}  \label{subch35}
%###################################################################################################

In harvesting processes, the grasp quality is one of the most important factors for production quality \cite{Miller}. In this scope, placed upon the harvesting robotic system’s manipulator, end-effectors have been in the researchers’ interest \cite{Rodriguez}. Being directly involved in the crop harvesting task, the design of these tools may require specific characteristics depending on the target crops type (e.g. shape, size). The required motion that an end-effector must perform in order to detach and harvest a fruit or a vegetable is usually considered within its design process, mostly attempting to mimic a human hand’s motion when grasping and picking a fruit from a tree while other end-effectors include a cutting device for crop detaching, simplifying the harvesting procedure. Various sensors may be integrated in the end-effectors, giving feedback that helps the overall robotic system acknowledge whether the harvesting target is picked up or not. 

 In this scope, a two bionic finger end-effector equipped with fiber sensors detecting the best position for grasping kiwifruits as well as pressure sensors to avoid damaging the fruits during grasping is proposed in \cite{Mu}. A multi-sensory configuration comprising distance, proximity, force and pressure sensors is considered in \cite{Liu3} for a universal end-effector for spherical fruits (e.g. apples, tomatoes, citrus). In particular, a two-fingered gripper singulates the fruit with a sunction pad while a laser device is responsible for cutting the fruit’s stem. Inspired by the operating principle of wire-stripping pliers, the end-effector’s upper and lower jaws cut and simultaneously hold the fruit by its peduncle in \cite{Jia}. A novel six-fingered end-effector equipped with three infrared (IR) sensors for active cutting position control is proposed in \cite{Xiong} for autonomous strawberries harvesting. Strawberry harvesting is also considered in \cite{Dimeas}, where a robotic gripper with three fingers and pressure profile sensors is developed in addition to a hierarchical control scheme, based on a fuzzy controller for the gripper’s force regulation and proper grasping criteria.

Besides the specifically pre-designed (non-configurable) end-effectors, gripper design automation is a new field of research that enables the existing equipment to be flexible and handle a larger range of tasks through reconfigurable grippers. A theoretical analysis taking into account the reconfigurability of the gripper is presented in \cite{Zhong}, where a balance between customization and simplicity is suggested. Most of the studies in the literature do not facilitate completely autonomous finger generation, rather following one of the three main approaches: modular design, re-configurable design and customized design. In modular design approaches \cite{Pedrazzoli}, there is a finger library where multiple types of finger geometries are stored. Then, based on a simplified geometry of the object to be grasped, the suitable pair of fingers is selected \cite{Pham}. A different technique has been introduced in \cite{Sanfilippo} where a grasping device adapts itself to the geometry of the object using a one DoF mechanism. These solutions while providing general and flexible solutions regardless the shape complexity, are computationally expensive because they iterate through each finger design for verification. In re-configurable designs \cite{Brown}, a gripper is designed with the possibility of re-configuring the fingers’ positions. The approach in \cite{Balan} employs a three-finger parallel jaw gripper where two of the fingers can change configuration while the third one is fixed. Such approaches while offering quick and simple solutions, require experts for the technical inputs of the algorithms. Finally, customized designs with custom fingers can be specifically generated for each object \cite{Pedrazzoli, Velasco, Velasco2, Honarpardaz}.

Grasp planning is defined as a search process of the possible locations for grasping an object, aimed at identifying closured grasps. The research area for this matter is wide and unstructured, mainly due to the fact that none of the proposed approaches has been widely implemented and considered the reference for comparison. In \cite{Sahbani} a clear overview of the methods and algorithms for grasp planning is presented and the authors divide the research done in this field into analytical approaches and data-driven (or knowledge-based) approaches.

Analytical approaches consist of methods that use geometric, kinematic and dynamic formulations in order to verify a grasp \cite{Bicchi}. The most basic approaches for grasp planning consider the minimum number of required contact points for force-closure of objects with polyhedral geometry \cite{JiaWei, Han, Mishra, Liu2}. The downside of the proposed multi-contact methods is the exhaustive search procedure. In heuristic methods, a large number of grasp positions are generated randomly \cite{Borst} by defining a set of rules that have been tested on the object’s model \cite{Miller2,Ding}. Then, unfeasible grasp candidates that do not fulfill the force-closure condition are filtered out. Although these approaches require low computational effort, they are usually limited to determining the local optima and don’t take every possible grasp position into account. Another limitation of multi-contact grasp methods that have been presented is that they are computationally efficient for very simple objects. Furthermore, methods for general object shape, model the object as a cloud of 3D points \cite{Ding2, Ding3, Eizicovits} or triangular mesh \cite{Carpin} and can be employed even for online grasp planning. Such methods \cite{Eizicovits} are employed in \cite{Hemming} for a robot harvesting sweet peppers using cameras on the end-effector and a 9 DoF manipulator. Although such algorithms can be applied to many different object shapes with multiple facets and reduced computational effort, they still suffer from local optima problems like the heuristic methods.

In data-driven approaches the grasp candidates are ranked based on a specific metric after being sampled for a particular object. Such methodologies are generally based on existing grasp experience, generated by simulation or on a real robot \cite{Sahbani}. A review of such approaches in \cite{Bohg} concludes that for each method the type and level of prior knowledge and the assumptions made about the objects being manipulated are the determining factor for their reliability. Grasping itself is highly dependent on the employed sensing and manipulation hardware, so the approaches cannot be compared with each other and benchmarked. In \cite{Kim}, automatic generation of possible grasp sets through a simulation-based method is described. This approach is different from most other studies as it focuses on the grasp process instead of the final grasp configuration. Furthermore, a date-driven method for industrial parallel-jaw grippers is proposed in \cite{Wolniakowski}. The algorithm starts with generating a grasp database by sampling on nearly parallel surfaces, and a dynamics based grasp simulator is used for evaluating the grasps. Finally, the suitable grasps are measured by three defined quality metrics. Such approaches, although computationally cheap, require a great deal of manual input and are not applicable for automated or online grasp planning.

%-----------------------------------------------------------------------------------------------------------------------------------------------------

\section{Related Research Projects}  \label{ch4}

\begin{figure}[h]
    \centering
    \includegraphics[width=.92\textwidth]{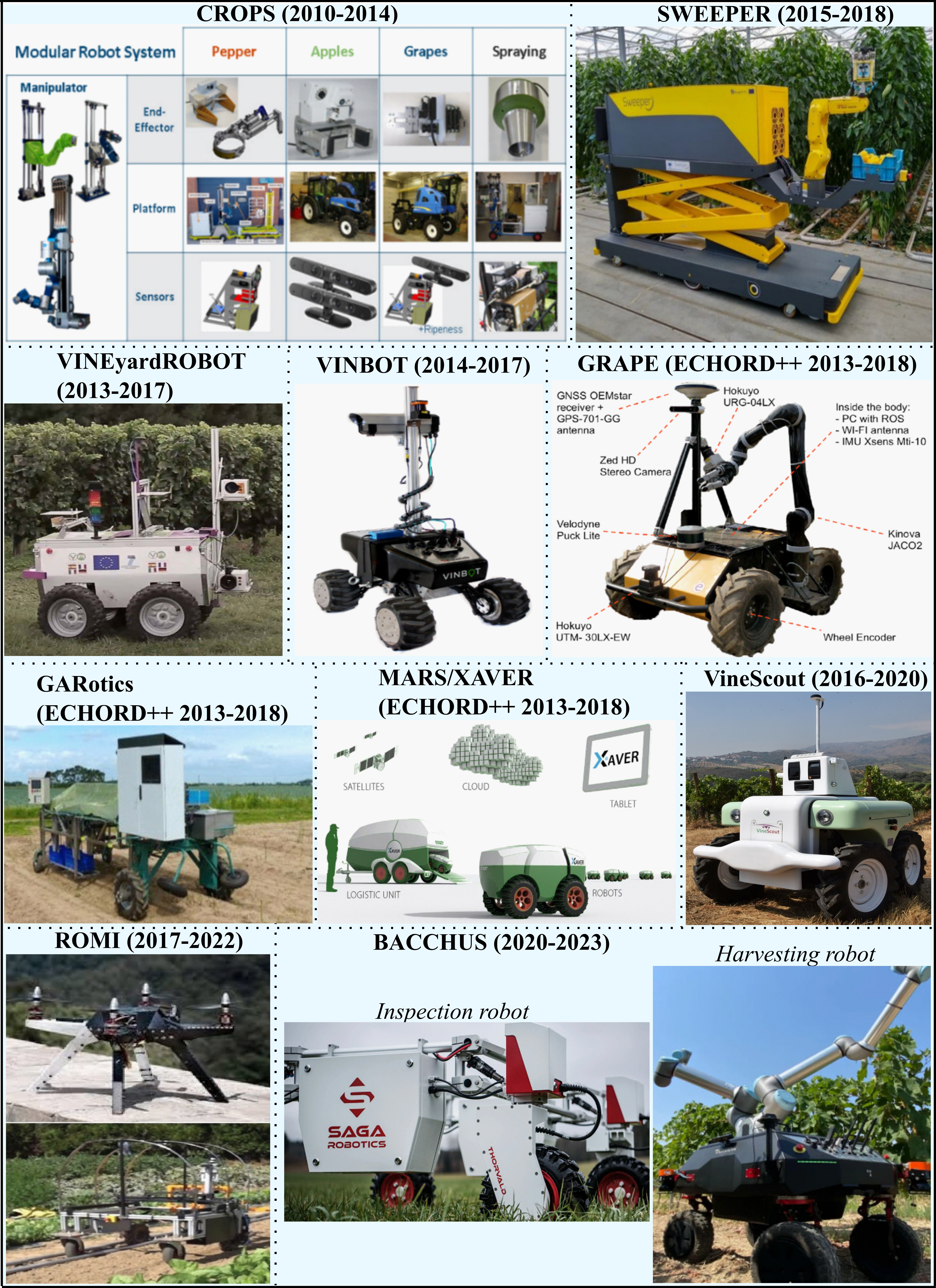}
    \caption{Research projects regarding various types of crops; subfigures source can be found in-text in each project's description 
    }
    \label{fig:proj}
\end{figure}

Several projects have been running during the last decade, regarding automated robotic agricultural applications, with Fig. \ref{fig:proj} presenting an overview of them. Mostly funded by the European Union (Seventh Framework Programme, Horizon2020\footnote{\href{https://cordis.europa.eu/programme/id/H2020-EC}{https://cordis.europa.eu/programme/id/H2020-EC} - Last accessed: 22-11-2022}), they involve the development of robotic systems whose primary goal is to facilitate and/or execute various agricultural tasks, such as harvesting, spraying, monitoring/phenotyping.

CROPS\footnote{\href{https://cordis.europa.eu/project/id/246252}{https://cordis.europa.eu/project/id/246252} \& \href{http://www.crops-robots.eu/}{http://www.crops-robots.eu/} - Last accessed: 22-11-2022} (Fig.\ref{fig:proj} - CROPS, subfigure taken from \cite{Best}) goal was to develop scientific know-how for a configurable, modular robotic system, including various modular tools (e.g. manipulators, sensors, sprayers, grippers) and adapting to several tasks. The robotic system developed within this project was capable of site-specific spraying and selective harvesting of sweet peppers and apples. Specifically, a 9 DoF manipulator was designed and tested for sweet-pepper and apple harvesting, as well as close range spraying in vineyards \cite{Best}. Different end-effectors and a prototype canopy sprayer were manufactured for each different application. Sensory system included color cameras, a multispectral system for fruit detection and discrimination between different elements of plants and a Time-Of-Flight (TOF) camera providing fast acquisition of distances, thus enabling localization of harvest targets \cite{Fernandez}. CROPS system was evaluated with regard to sweet pepper and apple harvesting in a commercial greenhouse and close range precision spraying in a vineyard \cite{Best}. Regarding sweet pepper harvesting, the developed system reported a 56-86\% detection rate and a relatively low harvest rate of 33\% with a high average harvest time of 94 seconds per fruit; however, the manipulator was intentionally moving with a low speed for safety reasons. Results for apple harvesting task, with special pruning firstly taking place on the crops, showed a 100\% detection rate and a 72\% harvest rate, requiring 15 to 30 seconds per apple. The system was successfully tested for close range target spraying in a greenhouse experiment with grapevines, reporting an 84\% attained pesticide reduction rate. An early performance evaluation concerning specifically the sweet pepper harvesting task under field conditions was also presented in \cite{Bac2}, reporting a difficulty in the successful picking of the harvest targets (6\% rate) and a clear improvement (from 6\% to 26\%) after simplifying the crop, that is removing fruit clusters, leaves and other obstacles. Lastly, grape harvesting was attempted during the project as well, however with no success since grape grasping was not successful most of the time with the gripper prototype being unable to cut the thick stalks of grape bunches even at maximum pressure.

SWEEPER\footnote{\href{https://cordis.europa.eu/project/id/644313}{https://cordis.europa.eu/project/id/644313} \& \href{http://www.sweeper-robot.eu/}{http://www.sweeper-robot.eu/} - Last accessed: 22-11-2022} - Sweet Pepper Harvesting Robot (Fig.\ref{fig:proj} - SWEEPER, subfigure taken from \href{http://www.sweeper-robot.eu/}{http://www.sweeper-robot.eu/}) was a H2020 project that built upon technologies, initially developed during the CROPS project. Its technical goal was to develop, test and validate a practical robotic harvesting solution for sweet peppers in real world condition. The developed robotic system involved a robotic vehicle with a single arm with a task-specific harvesting tool for the sweet pepper harvesting. For the identification of the fruits' location and the assessing of their maturity and quality, an RGB-D camera with custom made LED lightning fixtures was utilized, reporting simultaneously both color and depth information. The developed vision system included a Flash-No-Flash (FNF) controlled illumination acquisition protocol, facilitating target detection that is robust to ambient illumination effects \cite{Arad2}. The maturity and quality of the pepper was assessed based on color and shape features (perimeter/area ratio) measurements of the fruit. Before harvesting the fruits, a manual leaf-picking procedure was proposed, as in cases of image disturbances (e.g. leaves creating occlusions) the proposed system would not be able to identify the fruits. The idea of modifying target crops and their environment in order to make them “best fit” for robotic harvesting was considered during the overall project and methodologies to influence crop features were investigated and presented \cite{vanHerck} such as cultivation practices, pruning, climate control and artificial illumination. The developed robotic system was evaluated \cite{Arad} through the harvesting of 262 fruits along a 4 weeks period, where 104 fruits were part of a modified crop (low occlusion/cluster interference) and 159 of an unmodified one. The average achieved harvest time per fruit was 24 seconds, with logistics taking approximately half of that time. Harvest success rates were reported to be 61\% for best fit (modified) crop conditions and 18\% otherwise.

VINEyardROBOT\footnote{\href{https://cordis.europa.eu/project/id/610953}{https://cordis.europa.eu/project/id/610953} - Last accessed: 22-11-2022} or VINEROBOT (Fig.\ref{fig:proj} - VINEyardROBOT, subfigure taken from \cite{vinerob}) was an FP7 project, whose goal was the design and development of an agricultural robot for vineyard monitoring. Focusing on optimizing a vineyard’s management, VINEROBOT project’s basis consisted of building a ground robot/vehicle, endowed with artificial intelligence and several non-invasive sensing technologies. These technologies included fluorescence-based sensors, RGB machine vision, thermography, Visible (Vis) and Near-Infrared (NIR) spectroscopy and were combined with canopy images and data acquisition processes, executed in real time via customized algorithms. Monitoring the vineyard’s status and grapes composition e.g. amino acid concentration \cite{Fernandez2}, the proposed robotic system provided key information that would help the farmers in crop yield estimation \cite{Aquino}, vegetative growth and water status inspection \cite{Fernandez3}.

Focusing on helping the winegrowers to accurately assess their grapevines yield, VINBOT\footnote{\href{https://cordis.europa.eu/project/id/605630}{https://cordis.europa.eu/project/id/605630} - Last accessed: 22-11-2022} - VINyard roBOT project (Fig.\ref{fig:proj} - VINBOT, subfigure taken from \cite{Lopes}) aimed at developing an automatic robotic system that would monitor vineyards and provide useful relevant information and phyto-data in order to facilitate yield estimation and overall vineyard management. The developed robotic system involved an all-terrain autonomous mobile robotic platform with open source software. It was equipped with a set of sensors (e.g. color cameras, 2D laser rangefinders) capable of navigating through the field, while capturing and analyzing images and 3D data via cloud-based computing applications. The system’s objectives included the estimation of the amount of leaves and grapes (leaf-to-fruit ratio), canopy features like height, volume, and exposed leaf area, other phyto-data throughout the entire vineyard as well as the generation of online yield maps, that would greatly help winegrowers optimize their management strategies. Promising an autonomous monitoring of 168 hectares three times a year, an ability of climbing slopes up to $45^o$ and an electrical power autonomy of 8 hours per day, VINBOT aimed at representing a powerful precision viticulture cloud-computing agricultural tool, helping winegrowers produce higher quality wines. The evaluation of the developed system \cite{Lopes,Lopes2} reported good estimation results with acceptable accuracy regarding canopy features, with the overall grape yield however being underestimated due to various existing occlusions (bunch-on-bunch/leaf occlusions).

Part of ECHORD++\footnote{\href{https://cordis.europa.eu/project/id/601116}{https://cordis.europa.eu/project/id/601116} - Last accessed: 22-11-2022} FP7 project that funded various application-oriented research projects called “experiments”, GRAPE\footnote{\href{https://echord.eu/grape/}{https://echord.eu/grape/} \& \href{https://www.grape-project.eu}{https://www.grape-project.eu} - Last accessed: 22-11-2022} - Ground Robot for vineyArd monitoring and ProtEction (Fig.\ref{fig:proj} - GRAPE, subfigure taken from \cite{Astolfi}) aimed at creating a robotic system that would help with the vineyard monitoring and its overall protection. The project's partners included VITIROVER\footnote{\href{https://www.vitirover.fr/}{https://www.vitirover.fr/} - Last accessed: 22-11-2022}, a French company involved with vegetation robotic management and manufacturing of herds of mower-robots for soil grassing management. GRAPE’s main objective included the creation of enabling technologies that will allow precision agricultural practices, significantly reducing the negative environment impact of traditional farming (e.g. mechanical instead of chemical thinning, biological control instead of chemical pesticides). The developed robotic platform – UGV would detect plants, manipulate small objects and help with the vineyard’s health monitoring. A considered application of the system involved the automatic pheromone dispenser distribution for mating disruption (pest management) in vineyards \cite{Roure}. Specifically, the robot would monitor plants and their health through utilization of vision systems (multispectral camera, 2D laser sensor) and 3D real-time reconstruction; subsequently a biocontrol mechanism consisting of pheromone dispensers would be applied wherever needed for plague control. Additionally, a significant amount of research was conducted regarding the autonomous navigation of the robotic platform \cite{Astolfi}. With the vineyard environment presenting lots of challenges (e.g. weather conditions, soil and general terrain morphology, vegetation), GRAPE investigated utilization of various sensors and methodologies to facilitate the robot autonomous navigation through the field, including wheel encoders, GPS, LiDAR sensor, inertial measurement unit (IMU), Gmapping, Google’s Cartographer and others.

Aiming at facilitating asparagus harvesting and lowering its overall costs, GARotics\footnote{\href{https://echord.eu/garotics/index.php.html}{https://echord.eu/garotics/index.php.html} - Last accessed: 22-11-2022} (Fig.\ref{fig:proj} - GARotics, subfigure taken from \cite{garot}), also part of ECHORD++ FP7,  developed a prototype robotic system for selective green asparagus harvesting. The proposed system was able to (a) drive along an asparagus dam in the field, (b) detect asparagus stalks, identifying the ones ready for harvesting (involving three-dimensional-point cloud processing from RGBD image acquisition) and (c) perform harvesting without damaging the stalks, proposing a harvesting mechanism that utilizes a multi-tools solution for increased productivity. Initial field tests showed the harvester's applicability, reporting an average harvester's velocity of $0.2$ m/s with a mechanical harvesting cycle of approximately $2$ seconds, an average of $5$ harvested asparagus plants per meter with a two harvesting tools setup and about 90\% success rate in the harvesting process \cite{garot}.

MARS\footnote{\href{https://echord.eu/mars/index.php.html}{https://echord.eu/mars/index.php.html} - Last accessed: 22-11-2022} - Mobile Agricultural Robot Swarms project, an ECHORD++ FP7 "experiment" as well, has developed a robot system for high-precision planting, which has also been tested in the field. It uses small robots operating in swarms (6-12 units covering approximately 1 ha/h) and a cloud-based solution to plan, monitor and accurately document precise planting of corn. Satellite navigation and data management in the cloud allows operations to be conducted round the clock, with permanent access to all data. The position and planting time of each seed is accurately recorded, while planning for the required field, for seeds, seed patterns and density is carried out via an app. The intelligent OptiVisor algorithm plans the robot’s deployment based on the entered parameters, calculating the optimal paths for the units involved and the time required for completion of the job. Software updates for the system can be loaded “over the air”, just as a remote diagnostic can be run conveniently and in any location via the smart device. The latest version of the MARS robots was unveiled to the public for the first time at Agritechnica 2017, with the entire system, including small robots operating in swarms and a cloud-based system control now being operated under the product name “XAVER”\footnote{ \href{https://www.fendt.com/int/xaver}{https://www.fendt.com/int/xaver} - Last accessed: 22-11-2022} (Fig.\ref{fig:proj} - MARS(XAVER), subfigure taken from \href{https://www.fendt.com/int/xaver}{https://www.fendt.com/int/xaver}). 

VineScout\footnote{\href{https://cordis.europa.eu/project/id/737669}{https://cordis.europa.eu/project/id/737669} \& \href{http://vinescout.eu/web/}{http://vinescout.eu/web/} - Last accessed: 22-11-2022} (Fig.\ref{fig:proj} - VineScout, subfigure taken from \href{https://cordis.europa.eu/project/id/737669}{https://cordis.euro\\pa.eu/project/id/737669}) was the continuation of the completed project VineRobot. The project’s main goal was to industrialize, demonstrate, and take to market a field monitoring system (decision support system), embedded in a small-size and cost-efficient robot, regarding vineyards. Project’s objectives included optimization of the robotic system’s external design and internal electronics as well as the development of top performance navigation software and systems for protection and user friendliness. Conducted research within the project involved on-the-go utilization of Visible-Short Wave Near-Infrared spectroscopy \cite{Fernandez4} and hyperspectral imaging \cite{Gutierrez} as monitoring tools for grape composition (e.g. soluble solids, anthocyanins, polyphenols), thermal imaging for the assessment of vineyard’s water status \cite{Gutierrez22} as well as an investigation over the growing data-driven agriculture and the status of current advanced farm management systems \cite{Saiz}.

Ongoing project ROMI\footnote{\href{https://cordis.europa.eu/project/id/773875}{https://cordis.europa.eu/project/id/773875} \& \href{https://romi-project.eu/}{https://romi-project.eu/} - Last accessed: 22-11-2022} - RObotics for MIcrofarms (Fig.\ref{fig:proj} - ROMI, subfigures taken from \href{https://cordis.europa.eu/project/id/773875}{https://cordis.europa.eu/project/id/773875} - Results) develops an open and lightweight robotics platform for microfarms, found both in rural, peri-urban and urban areas, growing a large variety of crops (up to 100 different varieties of vegetables per year) on small surfaces (0.01 to 5 ha). ROMI's goal is to assist these farms in weed reduction and crop monitoring, reducing manual labor and increasing the productivity. Thanks to ROMI’s weeding land robot, farmers will save 25\% of their time, also acquiring detailed information on sample plants. This robot is coupled with a drone that acquires more global information at crop level, thus producing an integrated, multi-scale picture of the crop development that will help the farmer monitor the crops and increase efficiently harvesting. For this, ROMI aims to adapt and extend state-of-the-art land-based and air-borne monitoring tools to handle small fields with complex layouts and mixed crops. In addition to the ground and aerial robots, ROMI also introduces a 3D scanner for phenotyping in indoor and outdoor environments, combining an RGB camera with a powerful image processing pipeline for 3D plant representation and analysis.

With grape harvesting remaining an open and challenging task, new contributions can be significantly helpful, further advancing the use of automated agricultural technologies in this particular high-value field, i.e. vineyards. To this end, ongoing EU funded project BACCHUS\footnote{\href{https://cordis.europa.eu/project/id/871704}{https://cordis.europa.eu/project/id/871704} \& \href{https://bacchus-project.eu/}{https://bacchus-project.eu/} - Last accessed: 22-11-2022} (Fig.\ref{fig:proj} - BACCHUS, subfigures taken from \href{https://bacchus-project.eu/}{https://bacchus-project.eu/}) is developing an intelligent, mobile robotic system involving two independent and cooperative robots: (a) the inspection robot will navigate through a vineyard, inspecting and mapping crops, collecting various useful data, e.g. regarding their health and maturity, via an embedded sensorial system; (b) the harvesting platform will perform human-like harvesting with the needed finesse, introducing a bi-manual concept in contrast to most existing one-armed/uni-manual robotic solutions, thus offering additional manipulability, facilitating further the overall procedure. The envisioned solution aims at precision farming, applicable not only in vineyards but in other crops as well, incorporating smart use of robotics for advanced control, navigation/motion, adaptation and learning, towards increased farm productivity and reduced manual labor.

An overview of the above projects and their target functionalities is summarized in Table \ref{table:proj}, with some additional, technical information regarding their hardware/software characteristics as well.
\begin{table}[h!]
\centering
\begin{threeparttable}
\begin{tabular}{ |p{1.4cm}||p{5.6cm}|p{3.3cm}|  }                                                   \hline
\textbf{\makecell[l]{Projects}} 
        & \textbf{\makecell{Hardware/Software Specs}} 
        & \textbf{\makecell{Target functionalities}}                                             \\ \hline
                                                                                                    \hline 
                                                                                                  
\multirow{4}{1.4cm}{CROPS \cite{Best},\cite{Fernandez}, \cite{Bac2}}                                     & 
\multirow{4}{5.6cm}{9 DoF manipulator, modular end-effectors \& a prototype canopy sprayer, 
color cameras, multispectral system, time-of-flight camera, ROS software}                                &
\multirow{4}{3.3cm}{Sweet peppers \& apples harvesting, grapevine spraying, ripeness assessing}                                                                                                                           \\ & & \\ & & \\ & & \\ \hline

\multirow{5}{1.4cm}{SWEEPER \cite{Arad2},\cite{vanHerck}, \cite{Arad}}                                   & 
\multirow{5}{5.6cm}{Robotic vehicle \& a 6 DoF arm \& task-specific harvesting tool, 
RGB-D camera \& custom made LED lightning fixtures, Flash-No-Flash controlled illumination 
protocol, ROS software}                                                                                  &
\multirow{5}{3.3cm}{Sweet peppers harvesting with assessed maturity, crop/environment modification 
explored}                                                            \\ & & \\ & & \\ & & \\ & & \\ \hline

\multirow{5}{1.4cm}{\makecell[l]{VINEyard- \\ ROBOT} \cite{vinerob},\cite{Fernandez2},
\cite{Aquino},\cite{Fernandez3}}                                                                         & 
\multirow{5}{5.6cm}{Ground vehicle, fluorescence-based sensors, RGB machine vision, 
thermography, Visible and Near-Infrared spectroscopy}                                                    &
\multirow{5}{3.3cm}{Vineyard monitoring, amino acid concentration \& vegetative growth \& water status 
inspection, yield estimation}                                        \\ & & \\ & & \\ & & \\ & & \\ \hline

\multirow{4}{1.4cm}{VINBOT \cite{Lopes},\cite{Lopes2}}                                                   & 
\multirow{4}{5.6cm}{Mobile platform, Kinect v2 camera, 2D laser rangefinders, RTK-DGPS localization unit, 
ROS software}                                                                                            &
\multirow{4}{3.3cm}{Vineyard monitoring, yield estimation (leaf-to-fruit ratio), canopy information 
collection}                                                                 \\ & & \\ & & \\ & & \\ \hline

\multirow{4}{1.4cm}{GRAPE \cite{Astolfi},\cite{Roure}}                                                   & 
\multirow{4}{5.6cm}{Robotic platform \& 1 Kinova Jaco2\tnote{*} \ arm, multispectral camera, 2D laser 
sensor, GPS, LiDAR, inertial measurement unit, Gmapping, ROS software}                                   &
\multirow{4}{3.3cm}{Vineyard monitoring, pheromone dispenser distribution, health inspection}          
                                                                            \\ & & \\ & & \\ & & \\ \hline

\multirow{2}{1.4cm}{GARotics \cite{garot}}                                                               & 
\multirow{2}{5.6cm}{Wheeled platform, Microsoft Kinect v2 RGBD camera, 2 harvesting tools}               &
\multirow{2}{3.3cm}{Selective green asparagus harvesting}                                 \\ & & \\ \hline

\multirow{3}{1.4cm}{MARS/\\XAVER}                                                                        & 
\multirow{3}{5.6cm}{Small swarm robots, satellite navigation \& cloud-based data management, 
GPS-Real Time Kinematic technology}                                                                      &
\multirow{3}{3.3cm}{High-precision planting of corn seeds}                         \\ & & \\ & & \\ \hline

\multirow{4}{1.4cm}{VineScout \cite{Fernandez4},\cite{Gutierrez}, \cite{Gutierrez22},\cite{Saiz}}        & 
\multirow{4}{5.6cm}{Mobile platform, stereo camera, LiDAR \& ultrasound sensors, Visible-Short Wave 
Near-Infrared spectroscopy, hyperspectral imaging tools}                                                 &
\multirow{4}{3.3cm}{Vineyard monitoring, grape composition inspection (e.g. soluble solids, 
anthocyanins)}                                                              \\ & & \\ & & \\ & & \\ \hline

\multirow{3}{1.4cm}{ROMI}                                                                                & 
\multirow{3}{5.6cm}{Mobile platform \& drone, 3D scanner with RGB camera for 3D plant 
representation/analysis, C++ software}                                                                   &
\multirow{3}{3.3cm}{Weed reduction, crop monitoring, phenotyping}                  \\ & & \\ & & \\ \hline

\multirow{5}{1.4cm}{BACCHUS}                                                                             & 
\multirow{5}{5.6cm}{\textit{Inspection robot}: Thorvald \cite{Grimstad}, 
hyperspectral \& ZED2 RGB cameras, RTK GNSS Antennas, LiDAR. \textit{Harvesting robot}: Rb-Vogui\tnote{**}
\ , 2 UR10e arms, LiDAR, 4 ZED2 RGBD cameras, GPS. ROS software}                                         &
\multirow{5}{3.3cm}{Vineyards/grapes focus, crop monitoring \& mapping, assesed health \& maturity, bi-manual 
selective harvesting}                                                \\ & & \\ & & \\ & & \\ & & \\ \hline
\end{tabular}

\begin{tablenotes}\footnotesize
\item[*] \href{http://www.kinovarobotics.com/assistive-robotics/products/robot-arms}{http://www.kinovarobotics.com/assistive-robotics/products/robot-arms} - Last accessed: 22-11-2022
\item[**] \href{https://robotnik.eu/products/mobile-robots/rb-vogui-en/}{https://robotnik.eu/products/mobile-robots/rb-vogui-en/} - Last accessed: 22-11-2022 \\
\end{tablenotes}
\end{threeparttable}
\caption{Overview of projects and their technical specifications}
\label{table:proj}
\end{table}

%-----------------------------------------------------------------------------------------------------------------------------------------------------

\section{Related Commercial Products}  \label{ch5}

\begin{figure} [h!]
    \centering
    \includegraphics[width=.89\textwidth]{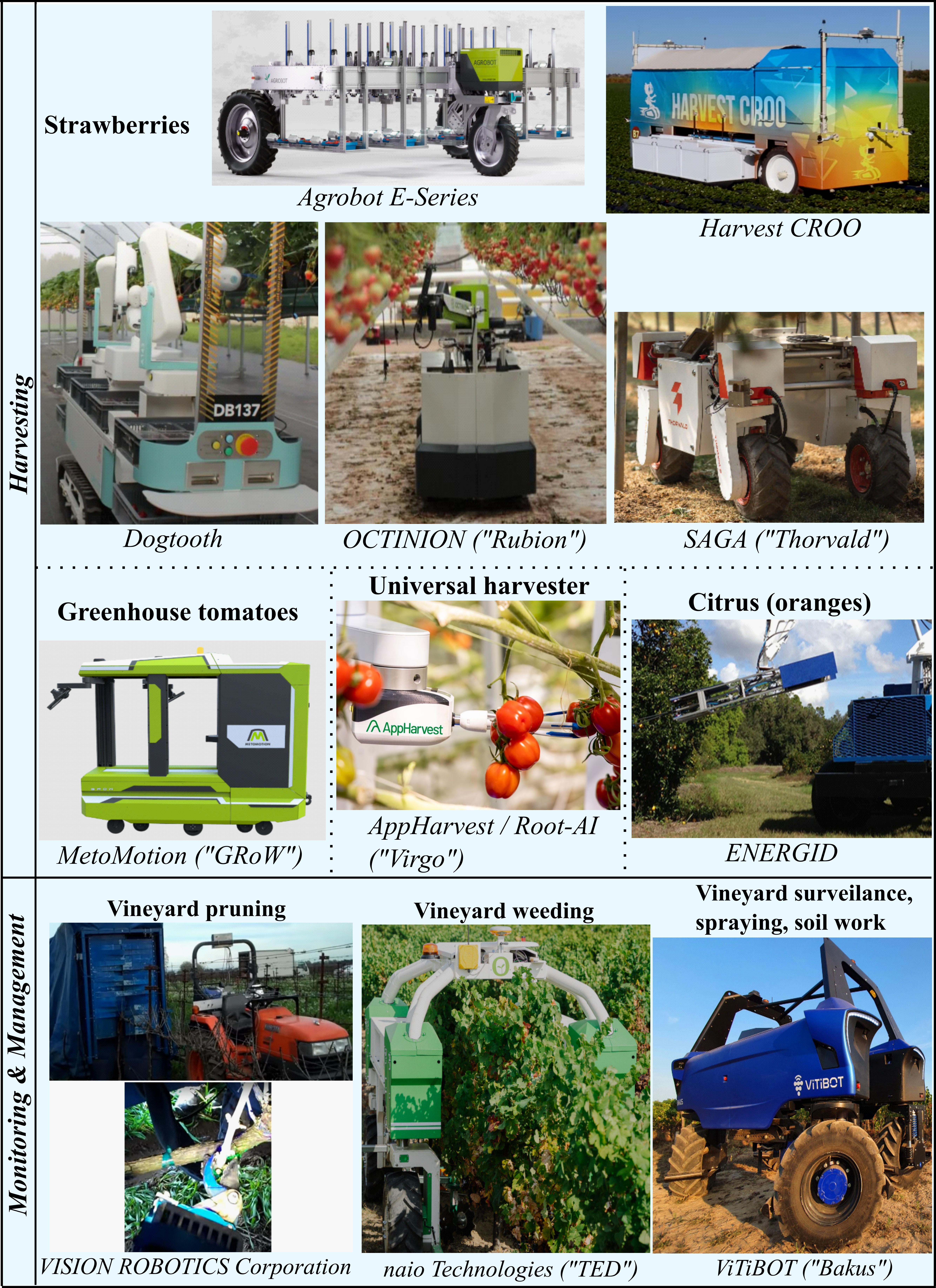}
    \caption{Commercial products regarding a variety of crops types and agricultural tasks; all subfigures taken from the products related websites, found in their respective citations/references}
    \label{fig:prod}
\end{figure}

In addition to research works and projects regarding automated harvesting and agriculture, several commercial products have been developed by various companies in the last decades, illustrated in Fig. \ref{fig:prod} (all subfigures taken from products’ websites that can be found in their respective citations/references).

Harvest CROO Robotics \cite{HarvestCROORobotics} was established in 2013, introducing an automated harvester for strawberries. Other strawberry robotic harvesters have been developed by Dogtooth \cite{Dogtooth} in the UK, Agrobot E-Series \cite{Agrobot} in Spain and OCTINION \cite{OCTINION} with their “Rubion” robot. In addition to their harvest robot, OCTINION has developed other robots for various agricultural activities besides harvesting (e.g. “Curion” for crop scouting, “Lumion” for fighting powdery mildew on strawberries, “Fluxion” for the logistic handling of storage crates). An autonomous strawberry harvester mounted on their robotic platform “Thorvald” is tested by SAGA Robotics \cite{SAGA} as well. The Thorvald robot, in particular Thorvald I \cite{Grimstad} and the further developed Thorvald II \cite{Grimstad2}, is an innovative module-based robot that allows for different robots to be build utilizing the same basic modules, hence addressing a variety of agricultural applications (e.g. crate transportation, UV-treatment, soil sampling, weeding).

The EU funded MetoMotion \cite{MetoMotion} developed the Greenhouse Robotic Worker - “GRoW” robot for selective robotic harvesting of greenhouse tomatoes, identifying and locating ripe fruits and performing robust damage free harvesting. Initially designed for greenhouse tomatoes as well, “Virgo” is a robotic universal harvester from Root-AI \cite{Root-AI}, now acquired by AppHarvest, that can be configured to identify and harvest multiple crops of varying sizes including tomatoes, peppers, cucumbers and strawberries; AppHarvest \cite{Appharv} focuses on controlled environment agriculture, involving smarter in-door farming as well as precise and eco-friendly growing (e.g. recycled rainwater, chemical pesticide-free fruits and vegetables). ENERGID \cite{ENERGID} has developed a robotic harvesting system for citrus reporting a 2 to 3 seconds harvest time per orange and a picking rate of 50\%, while Abundant Robotics was in the process of launching their commercial robotic apple harvester as well; however, it was unable to develop the market traction necessary to support its business during the COVID $19$ pandemic, resulting in its shut down\footnote{ \href{https://www.therobotreport.com/abundant-robotics-shuts-down-fruit-harvesting-business/}{https://www.therobotreport.com/abundant-robotics-shuts-down-fruit-harvesting-business/} - Last accessed: 22-11-2022}.

An interest has also been shown for grapevines, with VISION ROBOTICS Corporation’s intelligent autonomous grapevine pruner \cite{VISIONROBOTICS} and naio Technologies “TED” \cite{naio}. Utilizing GPS-RTK navigation and with an autonomy of up to 8 hours, “TED” is an autonomous weeding robot for vineyard maintenance. ViTiBOT \cite{ViTiBOT} is an industrial company located in France with a main objective of designing robotic platforms that facilitate eco-efficient work and vineyard surveillance. Their robot “Bakus” is equipped with a $360^o$ infrared vision system, capable of analyzing its environment thus avoiding obstacles. Reporting a 10-hour working autonomy during day or night and coping with slopes up to 45\%, it can provide several functionalities for vineyard maintenance, such as confined spraying, vine treatment and soil work. However, commercial selective robotic harvesters for grapes cannot be found. This is expected as the technologies to harvest such a delicate and geometrically complex fruit have apparently not reached a high technology readiness level.

%-----------------------------------------------------------------------------------------------------------------------------------------------------

\section{Conclusions}  \label{ch6}

A great amount of papers, projects and products may be found regarding the on-going research field of automated agriculture, focusing on autonomous harvesting but also on other tasks such as weeding, health monitoring, pruning, yield estimation and phenotyping. Apples, strawberries, tomatoes and sweet peppers occupy the majority of research interest, with grapes being targeted as well with respect to tasks other than harvesting, e.g. health monitoring, pheromone dispenser distribution, yield estimation and water status inspection. A uni-manual setup is adapted by most existing harvesting robotic solutions, with one arm both grasping and detaching the crop. BACCHUS is the only project addressing the challenging nature of the human-like, bi-manual approach (where one arm grasps the crop and the other cuts its stem). The additional manipulability of a bimanual solution facilitates the harvesting process,  clearly indicating an opportunity for new contributions by exploring further this approach. 
Development of specific functionalities typically required by an operating agricultural robot, has also been the focus of several works. Crop detection and navigation via vision have been shown to be  demanding research areas, with various reported issues open for addressing (e.g. crop occlusions by leaves/branches, crop’s color blending with its background, varying lighting conditions). 

With precision/automated agriculture being an open research field offering several challenges, further development of existing solutions as well as proposing new ones will be essential, towards making automated agriculture more efficient, precise and less time consuming. 

%-----------------------------------------------------------------------------------------------------------------------------------------------------

\section*{Statements and Declarations} 

\begin{itemize}

\item \textbf{Funding}: This research received funding from the European Community’s Framework Programme Horizon 2020 under grant agreement No 871704, project BACCHUS.

\item \textbf{Conflict of interest}: The authors declare that they have no conflict of interest.

\item \textbf{Author Contributions}: Conceptualization - Leonidas Droukas, Zoe Doulgeri; Literature search - Leonidas Droukas, Nikolaos L. Tsakiridis, Dimitra Triantafyllou, Ioannis Kleitsiotis and Dimitrios Kateris; Writing/original draft preparation - Leonidas Droukas, Nikolaos L. Tsakiridis, Dimitra Triantafyllou, Ioannis Kleitsiotis and Dimitrios Kateris; Writing/review and editing/critical revision - Zoe Doulgeri, Ioannis Mariolis, Dimitrios Giakoumis, Dimitrios Tzovaras and Dionysis Bochtis.

\item \textbf{Ethical approval}: Not applicable.

\item \textbf{Consent to participate}: Not applicable.

\item \textbf{Consent to publish}: Not applicable.

\item \textbf{Data and Code availability}: Not applicable.

\end{itemize}

%-----------------------------------------------------------------------------------------------------------------------------------------------------

\bibliography{main}

%-----------------------------------------------------------------------------------------------------------------------------------------------------

\end{document}